\documentclass{article}      
\usepackage{graphicx}		 
\usepackage{hyperref}
\usepackage{float}
\usepackage[]{algorithm2e}
\usepackage[
backend=biber,
style=numeric,
sorting=none
]{biblatex}
\usepackage[english]{babel}
\makeatletter
\newcommand\HUGE{\@setfontsize\Huge{50}{60}}
\makeatother

\title{Dynamic Neural Network Architectural and Topological Adaptation and Related Methods - \\  A Survey } 
\date{2021\\ May}
\author{Lorenz Kummer \thanks{Email lorenz.kummer@univie.ac.at}\\ Computer Science Department \thanks{Special thanks: Wilfried Gansterer}, University of Vienna}

\usepackage[margin=1.75in,heightrounded]{geometry}
\begin{document}

\maketitle

\begin{abstract}
Training and inference in deep neural networks (DNNs) has, due to a steady increase in architectural complexity and data set size, lead to the development of strategies for reducing time and space requirements of DNN training and inference, which is of particular importance in scenarios where training takes place in resource constrained computation environments or inference is part of a time critical application. In this survey, we aim to provide a general overview and categorization of state-of-the-art (SOTA) of techniques to reduced DNN training and inference time and space complexities with a particular focus on architectural adaptions.
\end{abstract}

\section{Introduction}
\label{sec:intro}
Recently published literature on strategies for minimizing DNN training and/or inference runtime and model size while maximizing accuracy can be broadly categorized into the following 4 categories based on the most important methodical concepts:
\begin{enumerate}
    \item Quantization (bit-width reduction of layers, i.e. weights, gradients, activations, sec. \ref{sec:quant}), to computationally more efficient floating-, fixed- or block-floating-point (an arithmetic approaching floating- on fixed-point hardware) \cite{wilkinson1994rounding} representations \cite{jacob2018quantization, yang2019quantization, zhang2018ECCV, sheng2018quantization, rajagopal2020multi, alistarh2017qsgd, alistarh2017qsgd_lg, loroch2017tensorquant, zhang2019qpytorch, achterhold2018variational, iandola2016squeezenet}. Quantization, in principle, can be applied either globally or on a per-layer basis, pre-training (apriori choice of a certain representation), intra-training (dynamic choice of representation during DNN training) or post-training (after training in float32, quantization to a lower representation is applied).
    \label{cat_quant}
    
    \item Adapting network architecture or topology (order, type, size or connections of layers, sec. \ref{sec:AA}) either via pruning of over-parameterized weights s.t. layers that do not contribute to the networks accuracy are either reduced in size, bypassed or removed completely and extending weights that under-parameterized \cite{gordon2018morphnet, singh2020woodfisher, han2015deep, iandola2016squeezenet, fang2020fast, hua2018channel, golub2018full, yang2020procrustes} or controlling network size via hyperparamaters \cite{sandler2018mobilenetv2, howard2017mobilenets, howard2019mobilenetv3, sinha2019thinmobilenet, chen2018hybridmobilenet}. Similar to quantization, architectural adaption can be applied either globally or on a per-layer basis, pre-training (pruning an architecture before training or re-training it), intra-training (dynamic pruning or extending of the DNNs weights or modifying its topology during training based on some criterion) or post-training (after training).
    \label{cat_arch}
    
    \item Introducing architectures obtaining their representational power (representational power based DNN taxonomy see \cite{khan2019survey}, sec. \ref{sec:new}) from entirely new, more efficient operations \cite{zhang2018shufflenet, ma2018shufflenet, iandola2016squeezenet}. Approaches employing this strategy usually require training from scratch. 
    \label{cat_new}
    
    \item Distributed (sec. \ref{sec:dist}) training and/or inference on large numbers of individually resource constrained devices \cite{ben2019demystifying, mcmahan2017federated, mcmahan2017federated2, wang2018deep, alistarh2017qsgd, alistarh2017qsgd_lg}
    \label{cat_dist}
\end{enumerate}
Strategies from categories \ref{cat_quant} and \ref{cat_arch} aim at optimizing DNN parameterization i.e. reducing over-parameterization either by keeping the number of parameters the same while reducing their numerical representations precision or by reducing the number of parameters while keeping their precision - both approaches lead to reduced model sizes and improved computational efficiency. Approaches falling into categories \ref{cat_new} and \ref{cat_dist} do not directly consider the number of model parameters or their numerical representation as relevant variables, but instead focus on more efficient use of computational resources for a given number of parameters and numerical representation. Besides a strategy-based super-categorization, we further sub-categorize literature in tab. \ref{tab:overview} dependent on the following parameters:
\begin{itemize}
    \item Whether abstract guarantees are provided by the authors (proof of convergence, time and space complexity, test error) 
    \item The data agnosticism of the algorithm (i.e. whether the content of the data is irrelevant to the algorithms function)
    \item Task type the authors used in their paper for demonstrating their approach (classification, object detection, speech recognition, semantic segmentation)
    \item The scenario where the approach claims an improvement over existing methods (training, inference, simulation - the last indicates the paper provides a simulation framework for e.g. simulating fixed-point quantization on floating-point hardware)
    \item Whether the approach is dynamic or static in the sense that it is computed during the target scenario or pre-computed
    \item Whether a publicly available code base exists that allows reproduction of the authors results, implemented either by the original authors or shipped as part of a machine learning framework e.g. \href{https://www.tensorflow.org/}{\underline{TensorFlow}}, \href{https://pytorch.org/}{\underline{PyTorch}}, \href{https://mxnet.apache.org/versions/1.8.0/}{\underline{MXNet}}.
\end{itemize}
Empirically measured speedups or testing errors were intentionally not included in tab. \ref{tab:overview} for they might not always be produced under comparable circumstances. If an approach matches the criteria for more than one category, it is listed in \ref{tab:overview} in all categories it fits into.

\subsection{Research and Selection of Literature}
\label{sec:intro:ssec:litsel}
The papers incorporated in our survey were researched through the search engines of Proceedings of Machine Learning Research (\href{http://proceedings.mlr.press/v119/}{\underline{PMLR}}), Neural Information Processing Systems (\href{https://nips.cc/}{\underline{NeurIPS}}), Institute of Electrical and Electronics Engineers (IEEE) \href{https://ieeexplore.ieee.org/Xplore/home.jsp}{\underline{Xplore}} as well as \href{https://scholar.google.com/}{\underline{Google Scholar}}. While we preferred works published in peer-reviewed journals or conference proceedings in the 2018-2021 timeframe, we included \href{https://arxiv.org/}{\underline{arXiv}} or earlier
papers as well if we deemed them particularly novel or creative. We selected papers based on the factors novelty, ingenuity,  citation count and reproduceability of results, i.e. the existence of a public code repository at e.g. \href{https://github.com/}{\underline{github}} or similar platforms or inclusion in a major machine learning framework such as \href{https://www.tensorflow.org/}{\underline{TensorFlow}}, \href{https://pytorch.org/}{\underline{PyTorch}} or \href{https://mxnet.apache.org/versions/1.8.0/}{\underline{MXNet}}.

\subsection{Related Surveys}
\label{sec:intro:ssec:relsurv}
The focus of other recent surveys in the field lies on parallel and distributed deep learning \cite{ben2019demystifying} whereby \cite{wang2018deep} specializes particularly on mobile applications. \cite{yu2018surveyquantized} treats theory and methods of quantization and discusses their merits and drawbacks while the complementary work \cite{tourad2020surveyfpgas} discusses frameworks and methods for executing DNNs on Field Programmable Gate Array (FPGA). DNN pruning techniques are surveyed and categorized by \cite{xu2020surveypruning}.
Recent developments in DNN architectures are described in \cite{khan2019survey}, which provides an overview not only over the SOTA but also the historical developments that lead there. The comparably new field of Quantum DNNs (QDNNs) is covered by \cite{li2020surveyquantumml, ramezani2020surveyquantumml}.
\\\\
Our survey differs from others by it's wider scope, incorporating works from different strategical categories, because we think that the most recent mixed-strategy approaches (sec. \ref{sec:mixed}) as well as the fact that quantization (sec. \ref{sec:quant}), DNN pruning (sec. \ref{sec:AA}) and the search for more efficient DNN architectures (sec. \ref{sec:new}) essentially treat the same issue - the systematic simplification of DNNs overly complex for the tasks they are intended to solve, leading to disproportional computation, memory and storage requirements - requires a broader perspective in order to provide a complete picture of the current SOTA. For brevity, our survey does not include recent advancements in QDNNs.

\section{Overview}
\label{sec:overv}
In this section, we will provide a categorization in tab. \ref{tab:overview} along the lines introduced in sec. \ref{sec:intro} as well as brief descriptions of mentioned approaches whereby particular attention will be paid to architectural adaption techniques.
\begin{table}[]
\begin{tabular}{|l|l|l|l|l|l|l|l|l|l|} \hline
\textbf{Paper} & \textbf{Cat.} & \textbf{Metric} & \textbf{Guar.} & \textbf{DA} & \textbf{Task} & \textbf{CA} & \textbf{Scenario} & \textbf{DS} & \textbf{Ref} \\  \hline
\cite{rajagopal2020multi} & \ref{cat_quant} & RT, TE, & N & Y & CL & \href{https://github.com/ICIdsl/muppet}{\underline{1}} & TR & D & \ref{sec:quant:ssec:quanttr} \\ \hline
\cite{jacob2018quantization, sheng2018quantization} & \ref{cat_quant} & RT, TE & N & Y & CL & \href{https://www.tensorflow.org/model_optimization/guide/quantization/training}{\underline{1}},\href{https://pytorch.org/docs/stable/quantization.html}{\underline{2}} & INF & S & \ref{sec:quant:ssec:quanti} \\ \hline
\cite{achterhold2018variational} & \ref{cat_quant} & TE & N & Y & CL & N & INF & S & \ref{sec:quant:ssec:quanti} \\ \hline
\cite{yang2019quantization} & \ref{cat_quant} & RT, TE & TC, SC & Y & CL, OB & \href{https://github.com/aliyun/alibabacloud-quantization-networks}{\underline{1}} & INF & S & \ref{sec:quant:ssec:quanti} \\ \hline
\cite{alistarh2017qsgd, alistarh2017qsgd_lg} & \ref{cat_quant}, \ref{cat_dist} & RT, TE & C, SC & Y & CL, SR & \href{https://github.com/scottjiao/Gradient-Compression-Methods}{\underline{1}} & TR & D & \ref{sec:mixed:ssec:dq} \\ \hline
\cite{loroch2017tensorquant, zhang2019qpytorch} & \ref{cat_quant} & TE & N & Y & CL & \href{https://github.com/cc-hpc-itwm/TensorQuant}{\underline{1}}, \href{https://github.com/Tiiiger/QPyTorch}{\underline{2}} & SIM & - & \ref{sec:quant:ssec:quantsim} \\ \hline
\cite{zhang2018ECCV} & \ref{cat_quant} & RT, TE & N & Y & CL & \href{https://github.com/Microsoft/LQ-Nets}{\underline{1}} & INF & S & \ref{sec:quant:ssec:quanti} \\ \hline
\cite{han2015deep} & \ref{cat_quant}, \ref{cat_arch} & RT, MS, TE & N & Y & CL & \href{https://github.com/mightydeveloper/Deep-Compression-PyTorch}{\underline{1}} & INF & S & \ref{sec:mixed:ssec:aaq} \\ \hline
\cite{iandola2016squeezenet} & \ref{cat_quant}, \ref{cat_arch}, \ref{cat_new} & RT, MS, TE & SC & Y & CL & \href{https://github.com/forresti/SqueezeNet}{\underline{1}} & TR, INF & S & \ref{sec:mixed:ssec:naq} \\ \hline
\cite{gordon2018morphnet} & \ref{cat_arch} & RT, MS, TE & N & Y & CL & \href{https://github.com/google-research/morph-net}{\underline{1}} & INF & D & \ref{sec:AAD:par:morph}\\ \hline
\cite{sandler2018mobilenetv2, howard2017mobilenets} & \ref{cat_arch} & RT, MS, TE & SC & Y & CL, OB & \href{https://github.com/tensorflow/models/tree/master/research/slim/nets/mobilenet}{\underline{1}}, \href{https://github.com/tensorflow/models/blob/master/research/slim/nets/mobilenet_v1.md}{\underline{2}} & TR, INF & S & \ref{sec:AAPreTA:par:mobnet}\\
\cite{ howard2019mobilenetv3}& & & & & SS & & & & \\ \hline
\cite{singh2020woodfisher} & \ref{cat_arch} & RT, MS, TE & TC, SC & Y & CL & \href{https://github.com/IST-DASLab/WoodFisher}{\underline{1}} & INF & S & \ref{sec:AAPostTA:par:wood} \\ \hline
\cite{fang2020fast} & \ref{cat_arch} & RT, MS, TE & N & Y & OB, SS & \href{https://github.com/JaminFong/FNA}{\underline{1}} & INF & S & \ref{sec:AAPostTA:par:FNA}\\ \hline
\cite{golub2018full} & \ref{cat_arch} & RT, MS, TE & N & Y & CL & N & TR, INF & D & \ref{sec:AAD:par:DP}\\ \hline
\cite{hua2018channel} & \ref{cat_arch} & RT, MS & N & N & CL & N & INF & D & \ref{sec:AAD:par:CGNets}\\
& & TE, EE & & & & & & & \\ \hline
\cite{yang2020procrustes} & \ref{cat_arch} & RT, MS, & N & Y & CL & N & TR, INF & D & \ref{sec:AAD:par:proc} \\ 
& & TE, EE & & & & & & & \\ \hline
\cite{zhang2018shufflenet, ma2018shufflenet} & \ref{cat_new} & RT, MS, TE & N & Y & CL, OB & \href{https://github.com/MG2033/ShuffleNet}{\underline{1}}, \href{https://github.com/pytorch/vision/blob/master/torchvision/models/shufflenetv2.py}{\underline{2}} & TR, INF & S & \ref{sec:new}\\ \hline
\cite{mcmahan2017federated, mcmahan2017federated2} & \ref{cat_dist} & RT, TE & N & Y & CL, SR & N & TR, INF & S & \ref{sec:dist}\\ \hline
\end{tabular}
\caption{\label{tab:overview} Overview of papers dependent on \textbf{Cat}egory, \textbf{Metric}, \textbf{Guar}antees, \textbf{D}ata \textbf{A}gnosticism, \textbf{Task}, \textbf{C}ode \textbf{A}vailability, \textbf{D}ynamic/\textbf{S}tatic and \textbf{Scen}ario. Papers sharing the same properties are summarized in the same row independent of academic relation. For a summary of the table, see sec. \ref{sec:disc}. Abbrevations: RT = \textbf{r}un\textbf{t}ime, MS = \textbf{m}odel \textbf{s}ize, TE = \textbf{t}est \textbf{e}rror, EE = \textbf{e}nergy \textbf{e}fficiency, C = \textbf{c}onvergence, T/SC = \textbf{t}ime/\textbf{s}pace \textbf{c}omplexity, CL = \textbf{cl}assification, OB = \textbf{o}bject \textbf{d}etection, SR = \textbf{s}peech \textbf{r}ecognition, SS = \textbf{s}emantic \textbf{s}egmentation, TR = \textbf{tr}aining, INF = \textbf{inf}erence, SIM = \textbf{sim}ulation, D = \textbf{d}ynamic, S = \textbf{s}tatic}
\end{table}

\subsection{Quantization}
\label{sec:quant}
\subsubsection{Quantized Inference} 
\label{sec:quant:ssec:quanti}
Minimizing inference accuracy degradation induced by quantizing weights and activations while leveraging associated performance increases can be achieved by incorporating (simulated) quantization into model training and training the model itself to compensate the introduced errors (Quantization Aware Training, QAT) \cite{jacob2018quantization, yang2019quantization}, by learning optimal quantization schemes through jointly training DNNs and associated quantizers (Learned Quantization Nets, LQ-Nets) \cite{zhang2018ECCV} or by using dedicated quantization friendly operations \cite{sheng2018quantization}. A notably different approach is taken by Variational Network Quantization (VNQ) \cite{achterhold2018variational} which uses variational Dropout training \cite{kingma2015variational} with a structured sparsity inducing prior \cite{neklyudov2017structured} to formulate post-training quantization as the variational inference problem searching the posterior optimizing the Kullback-Leibler-Divergence (KLD) \cite{kullback1951information}.
\subsubsection{Quantized Training} 
\label{sec:quant:ssec:quanttr}
For speeding up training on a single node via block-floating-point quantization, \cite{rajagopal2020multi} introduced a dynamic training quantization scheme (Multi Precision Policy Enforced Training, MuPPET) that tracks mini batch gradient diversity \cite{yin2018gradient} across epochs and decides if a precision switch is triggered for the next batch based on the violation of an empirically determined threshold . Due to a lack of dedicated fixed-point (i.e. FPGA) hardware, speed ups achieved by MuPPET were the product of simulations executed on floating-point hardware using NVIDIA CUTLASS \cite{nvidia2020cutlass}.
\subsubsection{Simulation} 
\label{sec:quant:ssec:quantsim}
For exploring the accuracy degradation induced by quantizations of weights, activations and gradients, \cite{loroch2017tensorquant} and \cite{zhang2019qpytorch} introduced the frameworks TensorQuant and QPyTorch capable of simulating on a float32 basis the most common quantizations for training and inference tasks. Both frameworks allow to freely choose exponent and mantissa for floating-point, word and fractional bit length for fixed-point and word length for block-floating-point representations as well as signed/unsigned representations. Since the quantizations remains a simulation, however, no actual speed ups can be achieved using these frameworks.

\subsection{Architectural and Topological Adaption}
\label{sec:AA}
\subsubsection{Pre-Training Adaption}
\label{sec:AAPreTA}
\paragraph{MobileNets} A representative of pre-training architectural adaption, MobileNets, first introduced by \cite{howard2017mobilenets}, are models that are based on depth-wise separable convolutions (highly efficient factorized convolutions, factorizing a standard convolution into a depthwise convolution and pointwise convolution \cite{sifre2014rigid}). Their width and resolution can be controlled by multipliers (i.e. hyperparameters) controlling layer in and out channels as well as input sizes, thus allowing architectural adaption. While the original paper introduces only a sequential model and does not offer a systematic approach how to find the best architecture for a certain use case, later extensions integrate non-sequential residual blocks \cite{sandler2018mobilenetv2} increasing representational power and employ Neural Architecture Search (NAS) algorithms \cite{howard2019mobilenetv3}. Numerous extensions of the popular MobileNet concept by researchers other than the original authors exist \cite{sinha2019thinmobilenet, chen2018hybridmobilenet}.
\label{sec:AAPreTA:par:mobnet}



\subsubsection{Post-Training Adaption}
\label{sec:AAPostTA}
\paragraph{WoodFisher} Based on the classic Optimal Brain Damage/Surgeon (OBD/S) \cite{lecun1990optimal, hassibi1993second} framework, WoodFisher \cite{singh2020woodfisher} uses second-order information in the form of efficiently approximated (Inverse-) Hessian's to determine the change in loss induced by the removal of one ore more parameters and prunes the network architecture accordingly. The approach is notable because not only does it not require retraining after pruning, but the authors also provide guarantees for time and space complexity.
\label{sec:AAPostTA:par:wood}
\paragraph{Fast Neural Network Adaption} Solving a problem different from reducing training or inference times but none the less interesting is Fast Neural Network Adaption (FNA) \cite{fang2020fast}: FNA uses depth-, width- and kernel level parameter remapping to map parameters from a pre-trained seed network to an arbitrary target network, thus allowing NAS to search for architectures optimized for other tasks (e.g. object detection, semantic segmentation) than the original seed network was (e.g. classification) without requiring retraining.
\label{sec:AAPostTA:par:FNA}

\subsubsection{Dynamic Adaption}
\label{AAD}
\paragraph{MorphNet} \cite{gordon2018morphnet} is an efficient resource-constrained structure learning algorithm based on the combination of width multipliers (as first introduced by MobileNet, see sec. \ref{sec:AAPreTA}) and sparsifying regularizers (e.g. L1 \cite{williams1995bayesian, ng2004feature, tibshirani1996regression}) to obtain particularly pruning-friendly weights tensors. MorphNet iteratively trains, shrinks, and expands a given DNN, finding each layers width multiplier s.t. a certain resource constraint (e.g. FLOPS or model size) is satisfied. While the authors are unable to proof convergence of their algorithm, they state that in their empirical evaluation, after 1 to 3 iterations, significant reductions in resource requirements where observed while maintaining iso-accuracy.
\label{sec:AAD:par:morph}
\paragraph{Dropback} introduced by \cite{golub2018full} is a DNN training algorithm which trains DNNs under a pruned weight budget. During training, Dropback tracks only the highest $k$ accumulated gradients ($k$ in this context refers to the weight budget) while untracked weights retain their initial values, thus reducing training times and memory accesses significantly. The actual model itself, however, is neither pruned in width nor depth and no advantage for inference is obtained.
\label{sec:AAD:par:DP}
\paragraph{Channel Gating} (CGNets) \cite{hua2018channel} is a dynamic inference topology pruning scheme that learns during training which regions in the features contribute least to the classification result and skips computations on a subset of these ineffective regions input channels. This is achieved by the introduction of learnable gate functions which, for each channel of a specific layer, learn if the channels output is zeroed out by ReLU or saturated to Sigmoid or TanH, allows bypassing of that channel in that layer.
\label{sec:AAD:par:CGNets}
\paragraph{Procrustes} \cite{yang2020procrustes} is an energy efficient sparse DNN training accelerator aimed at producing pruned models with iso-accuracy. Procrustes is based on above mentioned Dropback (sec. \ref{sec:AAD:par:DP}) and extends the concept by inducing sparsity through decaying the untracked weights initial values over the first $i$ iterations s.t. these weights can actually be pruned if they decay to zero, remedying one of Dropback's major drawbacks. Additionally, Procrustes avoids the need to sort all accumulated gradients by using dynamic quantile estimation to continuously track target sparsity.
\label{sec:AAD:par:proc}

\subsection{New Architectures}
\label{sec:new}
Instead of pruning, compressing or quantizing a pre-existing architecture, ShuffleNet \cite{zhang2018shufflenet} aims to be highly efficient per design: computationally expensive pointwise convolutions, used in SOTA architectures such as Xception \cite{chollet2017xception} or ResNext \cite{xie2017aggregated} but also the adaptive MobileNet or MorphNet architectures discussed in sec. \ref{sec:AAPreTA:par:mobnet} and \ref{sec:AAD:par:morph}, are replaced in ShuffleNet by less costly pointwise group convolutions. The introduced downside of grouping pointwise convolutions (outputs from a certain channel are derived only from a small fraction of input channels) are compensated by shuffling output channels s.t. the subsequent layers input channels receive features extracted from all other channels. The ShuffleNet concept is extended by ShuffleNetV2 \cite{ma2018shufflenet}, which introduces the concept of channel splitting to balance the numbers of groups and channels which were immutable in the original work.

\subsection{Distributed Training/Inference}
\label{sec:dist}
Proposing a new, completely decentralized view of DNN training, \cite{mcmahan2017federated, mcmahan2017federated2} introduce Federated Learning/Optimization (FL/O), an approach where no centralized learning takes place but instead client nodes (e.g. mobile devices) locally compute updates based on local data to a global model which is then averaged by a server node and shared among participating client nodes, thus performing a global update w.r.t. global data.


\subsection{Mixed}
\label{sec:mixed}
\subsubsection{New Architecture and Quantization}
\label{sec:mixed:ssec:naq}
By combining alternating pointwise and 3x3 convolutions into an architectural element called the Fire Module as well as post-training quantization, SqueezeNets \cite{iandola2016squeezenet} authors empirically show that their approach significantly reduces model size compared to other SOTA models while maintaining iso-accuracy. 

\subsubsection{Architectural Adaption and Quantization}
\label{sec:mixed:ssec:aaq}
Deep Compression \cite{han2015deep}, an earlier work, like SqueezeNet produces an efficient, quantized DNN as output but instead of beeing based on a new architectural element, Deep Compression applies pruning, quantization and Huffman coding \cite{huffman1952method} to a pre-trained network to achieve it's goal.

\subsubsection{Distributed Training and Quantization}
\label{sec:mixed:ssec:dq}
For training on multi-node environments (i.e. distributed), Quantized Stochastic Gradient Descent (QSGD) introduced by \cite{alistarh2017qsgd} incorporates a family of gradient compression schemes aimed at reducing inter node communication costs occurring during SGD's gradient updates, producing a significant speedup as well as convergence guarantees under standard assumptions (not included here for brevity, see \cite{alistarh2017qsgd_lg} for details).

\section{Discussion}
\label{sec:disc}
\subsection{Metrics}
\label{sec:disc:ssec:metrics}
As can be seen in tab. \ref{tab:overview}, a common ground all surveyed approaches share is they recognize test accuracy (expressed as test error by some works) as relevant metric. We conjecture this represents the common goal of iso-accuracy, i.e. any improvement in terms of time and space complexity must not come at the price of degraded test accuracy compared to more complex state-of-the art approaches. Second to test accuracy in terms of prevalence is runtime (expressed in time units, e.g. milliseconds, speedup or operation counts), which is recognized as relevant by all except 3 of the 24 surveyed papers. Those papers not recognizing runtime as relevant metric for their contribution were either quantization simulation frameworks (TensorQuant, QPytorch, sec. \ref{sec:quant:ssec:quantsim}) or did not disclose their reasons for why they decided against runtime as metric (VNQ, sec. \ref{sec:quant:ssec:quanti}). With regards to model size (expressed as number of parameters or bytes), notably no work falling solely into the quantization (sec. \ref{sec:intro} cat. \ref{cat_quant}) or distributed (sec. \ref{sec:intro} cat. \ref{cat_dist}) category uses this metric, however it is used by all works either introducing new architectures, adapting existing architectures (\ref{sec:intro} cat. \ref{cat_arch} and \ref{cat_new}) or mixing multiple approaches. While it is clear why work distribution strategies dont lead to decreased model sizes, a possible reason for the lack of reported model size reductions using quantization could be that quantization alone, while yielding improvements in runtime, does not result in a sufficient reduction in model size so that authors did not considers it's worthwhile reporting it - however, the verification of this conjecture is considered out of the scope of this survey. An interesting metric explicitly reported only by a 2 works (Procrustes and CGNets, sec. \ref{sec:AAD:par:proc} and \ref{sec:AAD:par:CGNets}) is energy efficiency: this is distinctive because energy efficiency (and indirectly carbon emissions) of deep learning algorithms have recently been subject of research and are suggested as key metric in evaluating deep learning models \cite{patterson2021carbon, anthony2020carbontracker}.
\subsection{Tasks}
The baseline task all surveyed works compete at is classification. In only 7 of 24 works, authors applied their approach to object detection, speech recognition or semantic segmentation tasks (namely QSGD (sec. \ref{sec:mixed:ssec:dq}), MobileNets (sec. \ref{sec:AAPreTA:par:mobnet}), FNA (sec. \ref{sec:AAPostTA:par:FNA}), ShuffleNets (sec. \ref{sec:new}), FL/O, \ref{sec:dist}) . There is no obvious connection between type of task chosen by the authors and strategical category, so we conjecture that authors might display the observed preference of classification task for reasons of simplicity.
\subsection{Guarantees, Data Agnosticism, Scenario}
Regarding abstract guarantees, only 5 of 24 works are able to provide guarantees w.r.t. convergence time and/or space complexity, namely QAT (sec. \ref{sec:quant:ssec:quanti}), QSGD (sec. \ref{sec:mixed}), MobileNets (sec. \ref{sec:AAPreTA:par:mobnet}) and WoodFisher (sec. \ref{sec:AAPostTA:par:wood}), and no work provides guarantees w.r.t. test error. The majority of authors rely solely on empirical evaluation of their works.
\\\\
Only CGNets (\ref{sec:AAD:par:CGNets}) considers input data in it's decision where to prune, which distinguishes it from the 23 other surveyed data agnostic approaches.
\\\\
There seems to be a focus on increasing inference performance in the surveyed literature, with 2 works focusing on training, 12 on inference and 10 offering improvements for both.

\subsection{Performance of Architectural Adaption Methods}
\label{sec:disc:ssec:perf}
In our work, we found that the most common experiment conducted by researchers working on DNN architectural adaption problems is classifying ImageNet \cite{russakovsky2015imagenet, deng2009imagenet} (seconded by classifying CIFAR-10/100 \cite{cstronoto20XXcifar}) and comparing the metrics test accuracy or error, Multiply-Add Operations (MAdds, 1 MAdd = 2 FLOPS)) and model size (usually given as number of parameters) with other SOTA approaches. Hence, we found this experimental setup most useful for comparing the performance of different approaches and summarized the authors results in tab. \ref{tab:perf}. This still remains only the smallest comparable subset of experiments however and should not be used to judge the contribution of an approach without considering its other properties (dynamic/static adaption, pre- or post-training, proven generalizability to other tasks such as object recognition or semantic segmentation, etc).

\subsubsection{The Best Solution?}
\label{sec:disc:ssec:perf:sssec:sol}
Within the limited expressiveness of the comparable experiments listed in tab. \ref{tab:perf} however, we found that Procrustes (sec. \ref{sec:AAD:par:proc}) outperforms all other works in terms of top-1 accuracy by model size ratio and top-1 accruracy to MAdds ratio, reporting a top-1 test accuracy of 71.13\% with only 75M MAdds and 0.35M weights when combined with MobileNetV2 (sec. \ref{sec:AAPreTA:par:mobnet}). The best top-1 accuracy by model compression ratio though is reached by WoodFisher (sec. \ref{sec:AAPostTA:par:wood}), which reports a top-1 test accuracy of 72.16\% after inducing 95\% sparsity in a ResNet50s weights, which under the assumption of a sparse weights storage format, translates to a reduction in model size by a factor of ~20. WoodFisher also outperforms all other approaches in terms of highest test accuracy: with the induction of 80\% sparsity, WoodFisher reaches 76.73\% accuracy and a model compression by a factor of 5. Unfortunately though the WoodFisher paper does not explicitly report MAdds or FLOPS but just speed-ups s.t. we cant compare the approach to others using this metric. At this point, we take into account that WoodFisher and Procrustes solve two similar but different problems: WoodFisher receives a pre-trained, un-pruned network and prunes it to obtain it's results while Procrustes already prunes during training, i.e. the latter is capable accelerating training and inference as well, which illustrates the limited usefulness of comparing experiments without taking algorithmic properties into account.

\begin{table}[]
\begin{tabular}{|l|l|l|l|l|l|}
\hline
\textbf{Approach} & \textbf{Top 1} & \textbf{MAdds} & \textbf{Size} & \textbf{Src} & \textbf{Ref} \\ \hline
MobileNetV1 (1.0) & 70.6 & 569M & 4.2M & \cite{howard2017mobilenets} & \ref{sec:AAPreTA:par:mobnet} \\ \hline
MobileNetV1 (0.75) & 68.4 & 325M & 2.6M & \cite{howard2017mobilenets} & \ref{sec:AAPreTA:par:mobnet} \\ \hline
MobileNetV1 (0.5) & 64.7 & 149M & 1.3M & \cite{howard2017mobilenets} & \ref{sec:AAPreTA:par:mobnet} \\ \hline
MobileNetV1 (0.25) & 50.6 & 41M & 0.5M & \cite{howard2017mobilenets} & \ref{sec:AAPreTA:par:mobnet} \\ \hline
GoogleNet (cite) & 69.7 & 1500M & 6.8M & \cite{howard2017mobilenets} &  \\ \hline
VGG16 \cite{simonyan2014very} & 71.5 & 15300M & 138M & \cite{howard2017mobilenets} &  \\ \hline
SqueezeNet & 57.5M & 1700M & 1.25M & \cite{howard2017mobilenets} & \ref{sec:mixed:ssec:naq}  \\ \hline
AlexNet \cite{krizhevsky2012imagenet}) & 57.2 & 720M & 60M & \cite{howard2017mobilenets} &  \\ \hline
MobileNetV2 (1.4) & 74.7 & 585M & 6.9M & \cite{sandler2018mobilenetv2} & \ref{sec:AAPreTA:par:mobnet} \\ \hline
MobileNetV2 (1.0) & 72.0 & 300M & 3.4M & \cite{sandler2018mobilenetv2} & \ref{sec:AAPreTA:par:mobnet} \\ \hline
ShuffleNetV1 (1.5) & 71.5 & 292M & 3.4M & \cite{sandler2018mobilenetv2} & \ref{sec:new}  \\ \hline
ShuffleNetV1 (2.0) & 73.7 & 524M & 5.4M & \cite{sandler2018mobilenetv2} &  \ref{sec:new} \\ \hline
MobileNetV3-L (1.0) & 75.2 & 219M & 5.4M & \cite{howard2019mobilenetv3} & \ref{sec:AAPreTA:par:mobnet} \\ \hline
MobileNetV3-L (0.75) & 73.3 & 155M & 3.0M & \cite{howard2019mobilenetv3} & \ref{sec:AAPreTA:par:mobnet} \\ \hline
MobileNetV3-S (1.0) & 67.4 & 66M & 2.9M & \cite{howard2019mobilenetv3} & \ref{sec:AAPreTA:par:mobnet} \\ \hline
MobileNetV3-S (0.75) & 65.4 & 44M & 2.4M & \cite{howard2019mobilenetv3} & \ref{sec:AAPreTA:par:mobnet} \\ \hline
InceptionV2 \cite{szegedy2016rethinking} & 74.1 & 5000M\cite{szegedy2016rethinking} & 25M \cite{szegedy2016rethinking} & \cite{gordon2018morphnet} & \\ \hline
MorphNet-InceptionV2 & 75.2 & 5000M & 25M & \cite{gordon2018morphnet} & \ref{sec:AAD:par:morph} \\ \hline
MobileNetV1 (0.5) & 57.1 & 149M & 1.3M & \cite{gordon2018morphnet} & \ref{sec:AAPreTA:par:mobnet} \\ \hline
MorphNet-MobileNetV1 (0.5) & 58.1 & 149M & 1.3M & \cite{gordon2018morphnet} & \ref{sec:AAD:par:morph} \\ \hline
MobileNetV1 (0.25) & 44.8 & 41M & 0.5M & \cite{gordon2018morphnet} & \ref{sec:AAPreTA:par:mobnet} \\ \hline
MorphNet-MobileNetV1 (0.25) & 45.9 & 41M & 0.5M & \cite{gordon2018morphnet} & \ref{sec:AAD:par:morph} \\ \hline
ResNet18 \cite{he2016deep} & 68.41 & 1800M\cite{yang2020procrustes} & 11M & \cite{golub2018full} & \\ \hline
Dropback-ResNet18 & 70.05 & ? & 2M & \cite{golub2018full} & \ref{sec:AAD:par:DP} \\ \hline
Dropback-ResNet18 & 67.99 & ? & 1M & \cite{golub2018full} &  \ref{sec:AAD:par:DP} \\ \hline
ResNet18 & 69.17 & 1800M & 11M & \cite{yang2020procrustes} &  \\ \hline
Procrustes-ResNet18 & 69.31 & 359M & 1M & \cite{yang2020procrustes} &  \ref{sec:AAD:par:proc} \\ \hline
MobileNetV2 (1.0) & 70.98 & 301M & 3.5M & \cite{yang2020procrustes} & \ref{sec:AAPreTA:par:mobnet} \\ \hline
Procrustes-MobileNetV2 (1.0) & 71.13 & 75M & 0.35M & \cite{yang2020procrustes} & \ref{sec:AAD:par:proc} \\ \hline
ResNet18 & 69.2 & 1800M & 11M & \cite{hua2018channel} & \\ \hline
CGNet-A-ResNet18 & 68.8 & 933M & ? & \cite{hua2018channel} & \ref{sec:AAD:par:CGNets} \\ \hline
CGNet-B-ResNet18 & 68.3 & 887M & ? & \cite{hua2018channel} & \ref{sec:AAD:par:CGNets} \\ \hline
MobileNetV1 (0.75) & 68.8 & 325M & 2.6M & \cite{hua2018channel} & \ref{sec:AAPreTA:par:mobnet} \\ \hline
CGNet-A-MobileNetV1 (0.75) & 68.2 & 173M & ? & \cite{hua2018channel} & \ref{sec:AAD:par:CGNets} \\ \hline
CGNet-B-MobileNetV1 (0.75) & 67.8 & 136M & ? & \cite{hua2018channel} & \ref{sec:AAD:par:CGNets} \\ \hline
ResNet50 & 77.01 & 3727M\cite{wang2019fully} & 20.69M\cite{wang2019fully} & \cite{singh2020woodfisher} & \\ \hline
WoodFisher-ResNet50 & 76.73 & ? & 4.14M & \cite{singh2020woodfisher} & \ref{sec:AAPostTA:par:wood} \\ \hline
WoodFisher-ResNet50 & 75.26 & ? & 2.07M & \cite{singh2020woodfisher} & \ref{sec:AAPostTA:par:wood} \\ \hline
WoodFisher-ResNet50 & 72.16 & ? & 1.03M & \cite{singh2020woodfisher} & \ref{sec:AAPostTA:par:wood} \\ \hline
\end{tabular}
\caption{\label{tab:perf} Experimental evaluation of architectural adaption approaches compared to 'classic' SOTA architectures (ResNet, AlexNet, VGG16, GoogleNet) by top-1 accuracy classifying ImageNet, MAdds (= Multiply-Add Operations, 1 MAdd = 2 FLOPS), Size (= Number of Parameters). MAdds and size are given in Millions.}
\end{table}







\section{Concluding Remarks}
We present a survey, which differs from other surveys in the field as outlined in sec. \ref{sec:intro:ssec:relsurv}, of carefully researched and selected (sec. \ref{sec:intro:ssec:litsel}) recently published SOTA methods in the field of dynamic DNN architectural and topological adaptation (sec. \ref{sec:AA}), quantized (sec. \ref{sec:quant}) and distributed (sec. \ref{sec:dist}) DNN training and inference as well as new, efficient architectures (sec. \ref{sec:new}). We further contribute a categorization (sec. \ref{sec:intro}) and overview (sec. \ref{sec:overv}) of these methods, describe the metrics (sec. \ref{sec:disc:ssec:metrics}) and experimental setups (sec. \ref{sec:disc:ssec:perf}) used commonly in the field as well as the problems faced when comparing different approaches (sec. \ref{sec:disc:ssec:perf}). Finally, we highlight open questions (sec. \ref{sec:disc:ssec:open}) we regard as relevant for their high potential to advance the SOTA in the field of deep learning.  

\section{Open Questions}
\label{sec:disc:ssec:open}
In DNN training, a DNN stores information on the shape of decision surface in it's weights. We found that no approach exists using a metric that directly measures the amount information lost (e.g. KLD) through weight pruning and we conjecture that the concept introduced by VNQ (sec. \ref{sec:quant:ssec:quanti}) could be extended to that purpose. We also point out that it has not yet been explored how quantized training impacts vanishing/exploding gradients, counter-strategies \cite{basodi2020gradient, chen2018gradnorm, salimans2016weight, klambauer2017self, glorot2010understanding, he2015delving, hanin2018start} and vice versa. It is furthermore unknown how the combination of various approaches performs. It might be worthwhile exploring how a MobileNet trained by Procrustes, pruned by WoodFisher and finally compressed and quantized by Deep Compression performs (sec. \ref{sec:AAPreTA:par:mobnet}, \ref{sec:AAD:par:proc}, \ref{sec:AAPostTA:par:wood}, \ref{sec:mixed:ssec:aaq}) and whether the results can be improved further by the most recent data augmentation techniques for semi-supervised and supervised deep learning tasks \cite{khosla2020supervised, chen2020simple}. 
\\\\
Given \cite{hong2019terminal} showed the practical feasibility of software-induced fault injection attacks on DNNs and the high vulnerability of DNNs to such attacks, another important question left untreated by all quantization related works in sec. \ref{sec:quant} and \ref{sec:mixed} is how these approaches hold up against dedicated attacks on quantized networks \cite{rakin2019bit, rakin2019bitimplementation} for they do not natively include defense mechanisms \cite{he2020defending, he2020defending2} against such attacks. Exemplary studies of the robustness of quantized neural networks against bit error attacks were recently conducted by \cite{giacobbe2020many, stutz2021bit}.
Another type of attack to which DNNs are particularly vulnerable are out-of-distribution (OOD) attacks \cite{augustin2020adversarial} which is why we deem it important to explore how architectural adaption or quantization affects a networks robustness to such attacks. Proofable guarantees w.r.t. to the detection of OOD attacks are provided by \cite{bitterwolf2020certifiable}.
\\\\
As discussed in sec. \ref{sec:disc}, only 5 of 24 surveyed works provide abstract guarantees regarding convergence, time or space complexity and not a single work provides any guarantees w.r.t. test error (i.e. network degradation induced by architecture modifications or quantization). We conjecture component-wise perturbation analysis \cite{highamperturbation} might lead to such guarantees as this technique has in similar scenarios not only been applied to study historical neural network architectures \cite{wangperturbation, meyerbaeseperturbation, zengperturbation} but also yielded results for simple modern architectures recently \cite{lenetperturbation}.
\\\\
Another question we deem worthy of further investigation is whether the interpretability \cite{zhang2018visual1, zhang2018visual2} of quantized, pruned or otherwise adapted DNNs changes w.r.t. to a base model and whether these changes, if present, can be quantified and lower or upper bounds for their magnitude be provided.


\printbibliography[type=inproceedings,title={References - Conference Papers}]
\printbibliography[type=article,title={References - Journal Papers}]
\printbibliography[type=inbook,title={References - Books}]
\printbibliography[type=misc,title={References - Other}]

\end{document}